\documentclass[acmtog]{acmart}

\usepackage{color,hyphenat,balance}
\usepackage[fleqn,tbtags]{mathtools}
\usepackage{autonum}
\usepackage{overpic}
\usepackage{wrapfig}
\usepackage{contour}
\usepackage{microtype}
\usepackage{multirow}
\usepackage{soul}
\usepackage{algorithm}
\usepackage{xspace}
\usepackage{algpseudocode}
\usepackage{diagbox}

\usepackage{mathtools}
\usepackage{lineno}

\setcopyright{acmcopyright}\acmJournal{TOG}
\acmYear{2022}\acmVolume{41}\acmNumber{6}\acmArticle{219}\acmMonth{12}
\acmDOI{10.1145/3550454.3555485}

\begin{document}

\title{Motion Guided Deep Dynamic 3D Garments}

\author{Meng Zhang}
\email{lynnzephyr@gmail.com}
\orcid{0000-0003-2384-0697}
\affiliation{%
 \institution{Nanjing University of Science and Technology}\country{China},\institution{University College London}
 \country{United Kingdom}}

\author{Duygu Ceylan}
\email{ceylan@adobe.com}
\orcid{0000-0001-6530-4556}
\affiliation{ \institution{Adobe Research} \country{United Kingdom}} 

\author{Niloy J. Mitra}
\email{n.mitra@cs.ucl.ac.uk}
\orcid{0000-0002-2597-0914}
\affiliation{ \institution{University College London and Adobe Research} \country{United Kingdom}} 





\renewcommand\shortauthors{Zhang M. et al}


\begin{abstract}

Realistic dynamic garments on animated characters have many AR/VR applications. While authoring such dynamic  garment geometry is still a challenging task, data-driven simulation provides an attractive alternative, especially if it can be controlled simply using the motion of the underlying character.
%
%
In this work, we focus on motion guided dynamic 3D garments, especially for loose garments. In a data-driven setup, we first learn a generative space of plausible garment geometries. 
Then, we   learn a mapping to this space  to capture the motion dependent dynamic deformations, conditioned on the previous state of the garment as well as its relative position with respect to the underlying body.  
%
Technically, we model garment dynamics, driven using the input character motion, by predicting per-frame local displacements in a canonical state of the garment that is enriched  with frame-dependent skinning weights to bring the garment to the global space. We resolve any remaining per-frame collisions by predicting residual local displacements. The resultant garment geometry is used as history to enable iterative roll-out prediction.
%
We demonstrate plausible generalization to unseen body shapes and motion inputs, and show improvements over multiple state-of-the-art alternatives. 
\textit{Code and data is released in \href{https://geometry.cs.ucl.ac.uk/projects/2022/MotionDeepGarment/}{https://geometry.cs.ucl.ac.uk/projects/2022/MotionDeepGarment/}}


\end{abstract}



\begin{teaserfigure}
  \includegraphics[width=\textwidth]{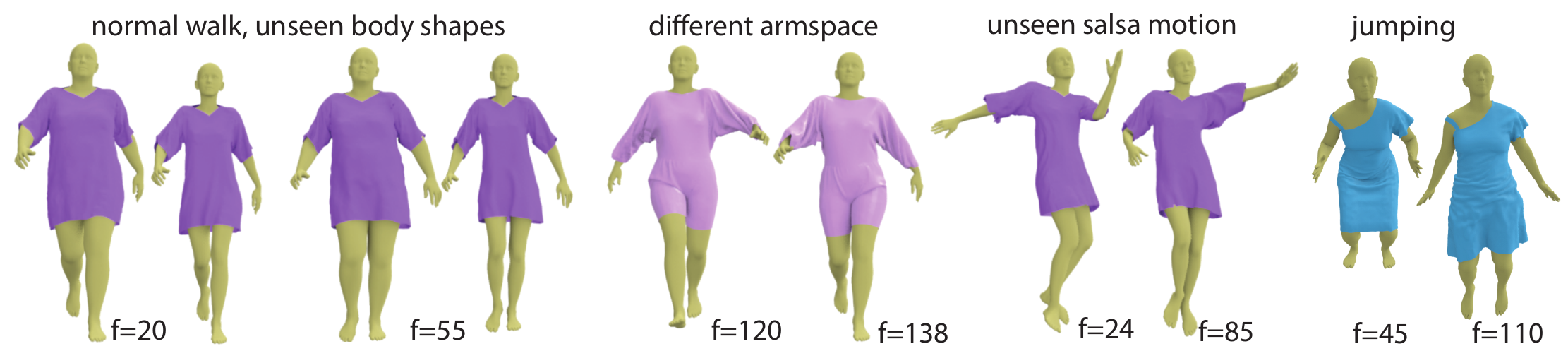}
  \caption{
  \textbf{Deep dynamic 3D garments. }
  Given a human body motion sequence and an initial garment state, we learn to generate realistic garment dynamics. Our network trained with a short normal walk sequence can generalize to unseen body shapes and out-of-training motion sequences (different armspace and unseen salsa motion). We can also train and test our network with more dynamic motion sequences such as jumping. We note that we train a garment-specific network for each garment type.} 
  \label{fig:teaser}
\end{teaserfigure}

\maketitle

\section{Introduction}

Authoring realistic garment dynamics, conditioned on human body motion, is an important problem with immediate applications in performance capture and digital retargeting for movies and games. %
Although physically-based simulations can increasingly model complex real-world human-garment interactions~\cite{Li2021CIPC}, they require access to complete material properties (e.g., spatially varying friction coefficient) and often require expert knowledge to setup.
Hence, there is a growing interest in the parallel approach of direct data-driven simulations~\cite{patel2020tailornet}. The latter approach is of particular interest as robust and high fidelity digital human capture~\cite{pons2017clothcap,CAPE:CVPR:20} becomes mainstream.

Any such (human) motion driven garment animation system should ideally satisfy the following: the output (garment) should be 
(i)~high fidelity and in full 3D, allowing seamless integration with existing production workflows (e.g., texturing and rendering); 
(ii)~free from temporal and/or spatial flickering; 
(iii)~respond plausibly to changes in target motion and human body proportions; 
(iv)~stable across long sequences, and (v)~free of interpenetration with the underlying human body.

Recent learning-based solutions geared towards motion driven garment animation largely focus on tight garments~\cite{alldieck2019learning}, where garment deformations can largely be approximated as constrained displacements with respect to the underlying body surface. In case of loose garments, a common approach is to focus on pose-conditioned garment deformations while ignoring the underlying body motion~\cite{patel2020tailornet,Bertiche2021,Bertiche2021ICCV}. Although showing plausible deformations at static poses, the predicted garments, specifically loose ones, still appear stiff and unrealistic across time. Hence, researchers have extended these setups to the temporal domain~\cite{santesteban2019learning,santesteban2021self,santesteban2022snug} while assuming the garment closely follows the underlying body motion by utilizing either fixed or constrained set of skinning weights. 
%
In contrast, as shown in Figure~\ref{fig:teaser}, we present a data-driven approach that responds to the underlying human motion to produce much more dynamic and plausible 3D geometry while generalizing to unseen body shapes and motion inputs. 

The space of body pose configurations and the corresponding garment geometries is large. When the dynamics (i.e., how fast a specific pose configuration is achieved) is also taken into account, the space grows even larger. Hence, learning such a space from a limited amount of training samples often results in severe overfitting with poor generalization to unseen motion. To overcome this challenge, we first learn a compact latent space of garment geometries that can act as a generative model for plausible garment deformations. We then learn a mapping from the previous states of the garment and its relative position with respect to the underlying body to this latent space to capture dynamic garment deformations. We decompose the garment deformation into local displacements predicted in the canonical state of the garment as well as a linear blend skinning that is driven by dynamically changing blending weights to transform the garment back to the posed space. Note that we do not have direct supervision for either the canonical space displacements or the dynamic blending weights. Instead, we supervise our approach using the final geometry of the garment in the posed space while enforcing regularization terms both on the canonical space deformations (e.g., edge length preservation, no body-garment collisions) as well as temporal latent mapping (e.g., in case of constant dynamics the mapping to the latent space of garment deformations is also constant). Finally, we represent the underlying body motion as a set of seed points independent of any specific body parameterization (e.g., SMPL~\cite{SMPL:2015}). This enables us to extend our method to handle multi-layer garments by treating the predictions of an inner layer garment as the body that derives an outer layer garment. Please note that the trained networks are garment-specific.
We evaluated our approach on a variety of garment types and tested generalization under out-of-training motion dynamics and body shapes. We compared our approach against three state-of-the-art alternatives~\cite{santesteban2021self,Bertiche2021,santesteban2022snug} and report that our method can capture more vivid and detailed deformations. 
In summary, our main contributions are as follows.
(i)~We present a novel learning setup that produces dynamic and plausible 3D garment geometry free of interpenetration with the body conditioned on input body motion sequence and a history of how the garment deforms. 
(ii)~We show stable rollout predictions by using the predictions of the method as input to capture the previous states of the garment. (iii)~{We show that by the first learning a generative space of plausible garment deformations, our method generalizes across unseen body shapes and motion sequences even when trained with a relatively short training sequence (e.g., 300 frames)}. Further, since our method does not depend on a specific body parameterization, we also demonstrate that it has the potential to be extended to handle multi-layer garments. 

\section{Related Work}

\paragraph{Physics-based simulation and data-driven approximations.} An accurate and principled approach to modeling dynamics of garments is to utilize physically-based simulation methods~\cite{Ko2005,Nealen2006,yu2019simulcap,tang2018cloth}. While being extremely accurate, such methods can be computationally expensive. Hence, there has been extensive work focusing on improving the efficiency~\cite{Weidner2018,liang2019differentiable,li2020p,wu2020safe}. However, the computational efficiency and robustness still heavily depend on the geometric complexity of the garments and having access to physical parameter values. 

To reduce computational efficiency some methods have focused on approximating the physically based simulation process by predicting a high resolution garment mesh from a coarse one. Typical approaches to achieve this goal include constraint-based optimization methods~\cite{Muller2010,Rohmer2010,Gillette2015} or data-driven methods. Among the data-driven approaches, while some learn a mapping from the coarse garment mesh to fine-scale displacements~\cite{Feng2010,zurdo2012animating}, others learn upsampling operators~\cite{Kavan2011}. High resolution garment deformations can also be obtained by interpolating and blending examples from a database~\cite{Wang2010,xu2014sensitivity}, or by operating at a reduced subspace of garment deformations~\cite{guan2012drape,hahn2014subspace}.

\begin{table}[t!]
\caption{We classify the related works that also utilize a skinning method with respect to three factors: static (conditioned on pose only) or dynamic deformations, use of fixed or dynamic skinning weights, and garment-body collision handling (if no, only a post optimization is used). }
\label{table:classify}\small
\begin{tabular}{r||c|c|c}
& deformations & skinning weights & collisions \\
\hline \hline
Santesteban et al.~\shortcite{santesteban2019learning} & dynamic & fixed & yes  \\
\hline 
Gundogdu et al.~\shortcite{gundogdu2019garnet} & static & fixed & yes  \\
\hline
Patel et al.~\shortcite{patel2020tailornet} & static & fixed & no  \\
\hline 
Santesteban et al.~\shortcite{santesteban2021self} & dynamic & dynamic & yes  \\
\hline 
Tiwari et al.~\shortcite{Tiwari_2021_ICCV} & static & fixed & yes \\
\hline
Bertiche et al.~\shortcite{Bertiche2021ICCV} & static & dynamic & yes  \\
\hline 
Bertiche et al.~\shortcite{Bertiche2021} & static & dynamic & yes  \\
\hline 
Santesteban et al.~\shortcite{santesteban2022snug} & dynamic & fixed & yes  \\
\hline 
ours & dynamic & dynamic & yes 
\end{tabular}
\end{table}

\paragraph{Learning-based garment deformations.} Recently, deep learning based methods have become popular alternatives to approximate physically-based simulations. 
Wang et al.~\shortcite{wang2018learning} predict PCA coefficients of draped garments on different body shapes in a canonical pose from sketch input. The earlier work of Yang et al.~\shortcite{yang2018analyzing} first extracts a garment layer from 3D dense scans of humans with clothing and then learns a reduced PCA based garment deformation space. Holden et al.~\shortcite{holden2019subspace} learn a mapping between the parameters of an external force such as the character motion and a reduced PCA-based representation of a cloth. In such setups, the dimensions of the PCA space has a direct impact on the geometric details that can be recovered. Increasing the PCA dimensions, however, increases the computational cost resulting in a trade-off. Wang et al.~\shortcite{wang2019learning} represent such a reduced space using a neural network. Assuming the state of the garment is provided at selected keyframes, the method predicts the shape of the garment conditioned on the body motion. While these methods learn a reduced space of the final garment deformations, we utilize a regularized autoencoder to learn a generative space of garment deformations in the canonical pose and rely on a skinning function to compute the final garment geometry. This decoupling enables our method to capture more detailed deformations.

A large body of work has focused on predicting deformations of tight clothing as constrained displacements with respect to the underlying body~\cite{alldieck2019learning,bhatnagar2019multi,pons2017clothcap,10.1111/cgf.14108,ma2020learning}. In order to provide a more general approach for handling loose garments,  methods have explored implicit and point based representations to predict the shape of garments under a specific body pose~\cite{Saito:CVPR:2021,Ma:CVPR:2021,ma2021power,tiwari21neuralgif}. These approaches, however, do not model the garments separately. Specifically, garment geometry is represented in a coupled manner to the underlying naked body limiting the scope of such methods. SMPLicit~\cite{corona2021smplicit} represents garments using an unsigned distance field with respect to the body. However, it models only body pose and focuses on predicting the overall shape of the garment instead of detailed deformations. In another line of work, researchers model, in a two-stage process, coarse deformations and fine details. Lahner et al.~\shortcite{lahner2018deepwrinkles} first uses an LSTM type of architecture to predict garment deformations in a reduced linear subspace followed by detail enhancement; while Zhang et al.~\shortcite{Zhang2020} cast detail enhancement as a style transfer task. However, they rely on access to  coarse simulation to first capture the overall deformation of the garment.

A more popular recent trend is to model garment deformations by utilizing a skinning based model, one we also adopt. More specifically, given a garment template mesh at rest pose and a skinning function that relates the garment geometry to the underlying body motion, several works focus on predicting additional displacements to capture detailed garment deformations. While Gundogdu et al.~\shortcite{gundogdu2019garnet} predict residual displacements after skinning, several works focus on predicting the deformation of the garment in a canonical pose followed by the skinning. We characterize these works which are closest to ours based on several factors as shown in Table~\ref{table:classify}. In the first group, \cite{patel20tailornet,Tiwari_2021_ICCV,Bertiche2021ICCV,Bertiche2021} model deformations conditioned only on the body pose, ignoring effects of body motion. The recent works of Santesteban et al.~\shortcite{santesteban2019learning,santesteban2021self,santesteban2022snug} use a GRU-based architecture to model the dynamics. In contrast, we explicitly access information about both the garment and body velocity and acceleration to learn motion-dependent deformations, which are regularly observed in loose garments. Another important aspect is the choice of the skinning function. In a simplified setup, some works assume a fixed set of skinning weights computed for the garment geometry in a canonical pose based on the proximity of the garment to the underlying body~\cite{gundogdu2019garnet,santesteban2019learning,patel20tailornet,Bertiche2021,santesteban2022snug}. 
Santesteban et al.~\shortcite{santesteban2021self} relax this constraint and present a \emph{diffused human model} by smoothly diffusing skinning parameters to any 3D point around the body. Bertiche et al.~\shortcite{Bertiche2021ICCV,Bertiche2021} jointly predict per-vertex deformations and blending weights. Inspired by recent work~\cite{yang2021lasr,yang2022banmo}, we utilize 3D Gaussian ellipsoids that move along with the bones to define dynamic skinning weights. We find that this provides a good trade off compared to a diffused human model and unconstrained per-vertex weights (see Section~\ref{sec:exp}). Last but not least, another important differentiating factor across related work is the treatment of collisions between the body and the garment. While some methods do not explicitly use any collision term during training~\cite{patel2020tailornet}, others define a collision loss~\cite{gundogdu2019garnet,Bertiche2021,Bertiche2021,santesteban2022snug} similar to our work. Similar to related work, we observe that this term is not sufficient to guarantee collision-free deformations, especially for complex garments. While a per-frame post optimization~\cite{patel2020tailornet} to push colliding vertices outside the body is a common approach, we instead introduce a test-time optimization scheme where we only need to resolve collisions on a sparse set of frames (see Section~\ref{sec:exp}). Finally, the work of Santesteban et al.~\shortcite{santesteban2021self} handles body-garment collisions by first optimizing for collision free unposed deformed garments using their diffused human model. By supervising their method with such unposed garments, they show effective handling of collisions. However, as we show in our comparisons, the diffuse human model assumption limits the range of dynamic deformations that can be predicted (see Section~\ref{sec:exp}).

More recently, Pfaff et al.~\shortcite{pfaff2021learning} develop a very interesting graph-based network to learn mesh-based simulations. While showing impressive results, this method operates on a complete graph
composed of all the mesh vertices. Hence, it is not straightforward
to extend it to handle complex garment geometries and interactions
between the garment and the underlying body. 

\section{Overview}
Given a 3D character body $B$ and a target garment geometry $G$, simulated under a certain motion sequence at training time, our method predicts how the garment would deform over a target body, potentially with a new set of body shape parameter, under a new (i.e., unseen) motion sequence. We assume that the geometry of the garment at time $t$, i.e., $G_t$, depends on the state of the garment in the previous step, i.e., its geometry $G_{t-1}$, velocity (field) $V_{t-1}$, and acceleration (field)  $\dot{V}_{t-1}$. Further, the garment deformation is also influenced by the interaction between the body and the garment. Hence, our goal is to design a neural network that approximates the current state of the garment as a function of the previous state of the garment and the current state of the body.

\begin{figure}[b!]
  \includegraphics[width=\columnwidth]{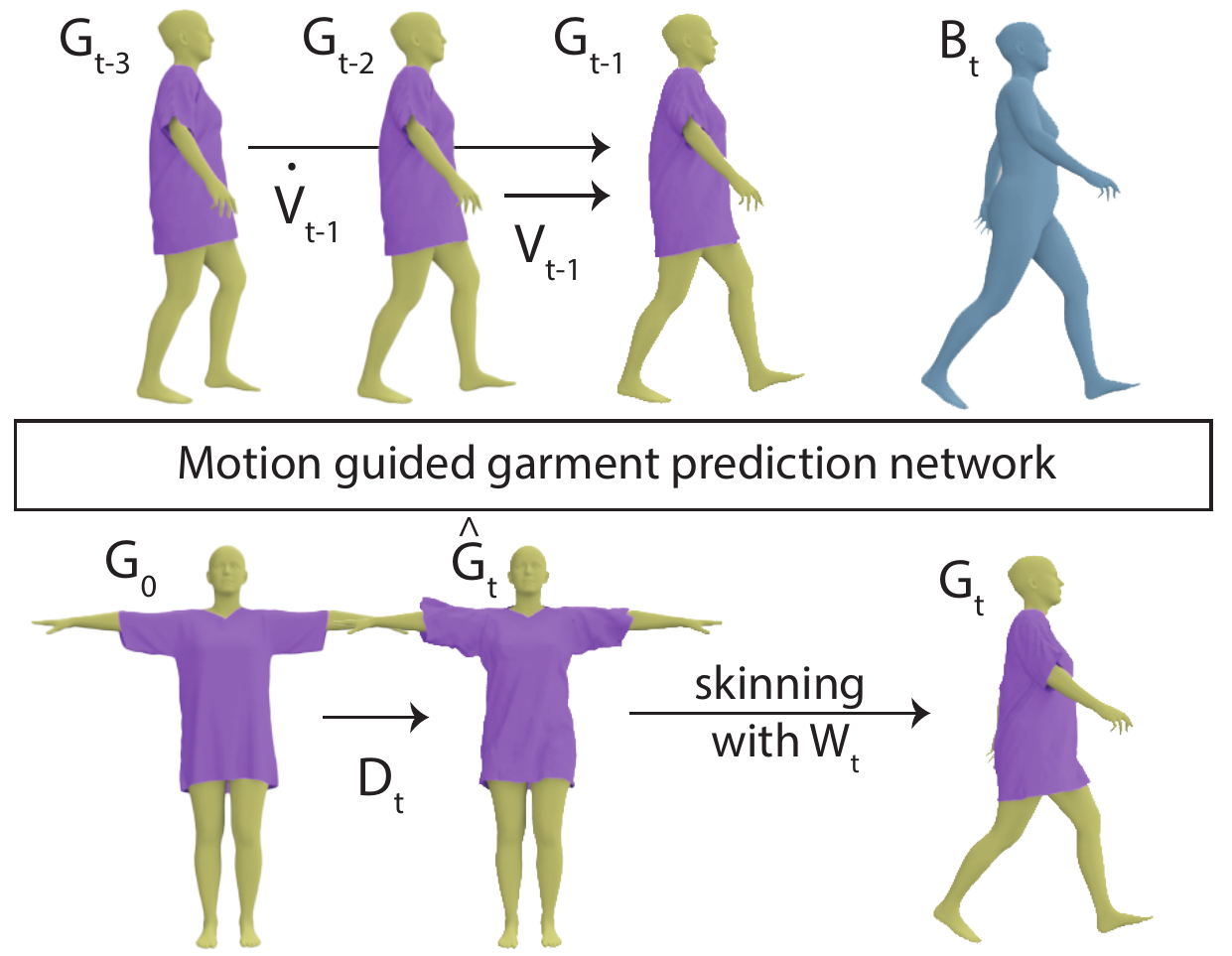}
  \caption{
  \textbf{Method overview. }
  We present a motion guided 3D garment prediction network that takes as input the previous state of the garment (i.e., the garment geometry, $G_{t-1}$, velocity, $V_{t-1}$, and acceleration, $\dot{V}_{t-1}$ at time $t-1$) and the current body (i.e., $B_{t}$, the body geometry at time $t$) and predicts $G_t$, the garment geometry at time $t$. Garment deformation is factorized into local displacements $D_t$ with respect to the canonical garment state, $G_0$, and linear skinning driven by $W_t$, dynamic blending weights.}
  \label{fig:overview}
\end{figure}

We aim to learn a generalizable model that can approximate the large space of body pose configurations and the corresponding garment geometries along with dynamics (i.e., how fast a specific pose configuration is achieved). To achieve this goal, we learn a compact latent space that acts as a generative model of plausible deformed garment geometries. We adapt the concept of \emph{pose space deformations}~\cite{Lewis2000} and decompose the garment deformation into (i)~local displacements $D_t$ that are represented with respect to a canonical rest shape of the garment and (ii)~global deformation that is computed by linear skinning based on the underlying body motion. Instead of assigning a fixed set of skinning weights to the garment vertices, we predict per-frame dynamic weights, $W_t$, which prove effective in modeling the deformation of loose garments. Since we do not have access to either ground truth canonical space displacements or dynamic blending weights, learning a direct mapping from previous garment and body states to $(D_t,W_t)$ is challenging. Hence, we first train a regularized autoencoder that maps a static deformed garment geometry to a generative latent space that can then be used to predict plausible $(D_t,W_t)$ configurations (Section~\ref{subsec:garment geometry prediction}). We represent the input garment geometry with a descriptor $P_{t}$ that encodes the relative vertex positions of $G_{t}$ with respect to the body $B_{t}$. This relative descriptor enables our network to generalize, at test time, to unseen body shapes and pose configurations. Once we learn a generative mapping from $P_{t}$ to $(D_t,W_t)$, in a subsequent stage, we train a dynamic aware encoder  (Section~\ref{sec:dynamic-aware encoding}). The dynamic encoder maps the previous state of the garment (i.e., garment geometry, velocity, and acceleration) along with the body interaction to a latent code in the generative space, which is then used to decode $(D_t,W_t)$ for the current frame.
%
Figure~\ref{fig:overview} provides an overview of 
our approach.

\section{Generative Space of Garment Deformations}

 \subsection{Garment deformation model} \label{Sec: repre_garment_deformation}
 While it is possible to represent $G_t$ with absolute coordinates in the world space, a common practice in articulated body animation is to adapt pose space deformation~\cite{Lewis2000}. Specifically, given the geometry of the garment in a canonical reference state $G_0$, we represent $G_t$ as local displacements $D_t$ with respect to $G_0$, and a global deformation computed by linear skinning based on the body motion driven by the \textit{time-varying} blending weights $W_t$.
 
\begin{figure*}[t!]
  \begin{overpic}[width=\textwidth]{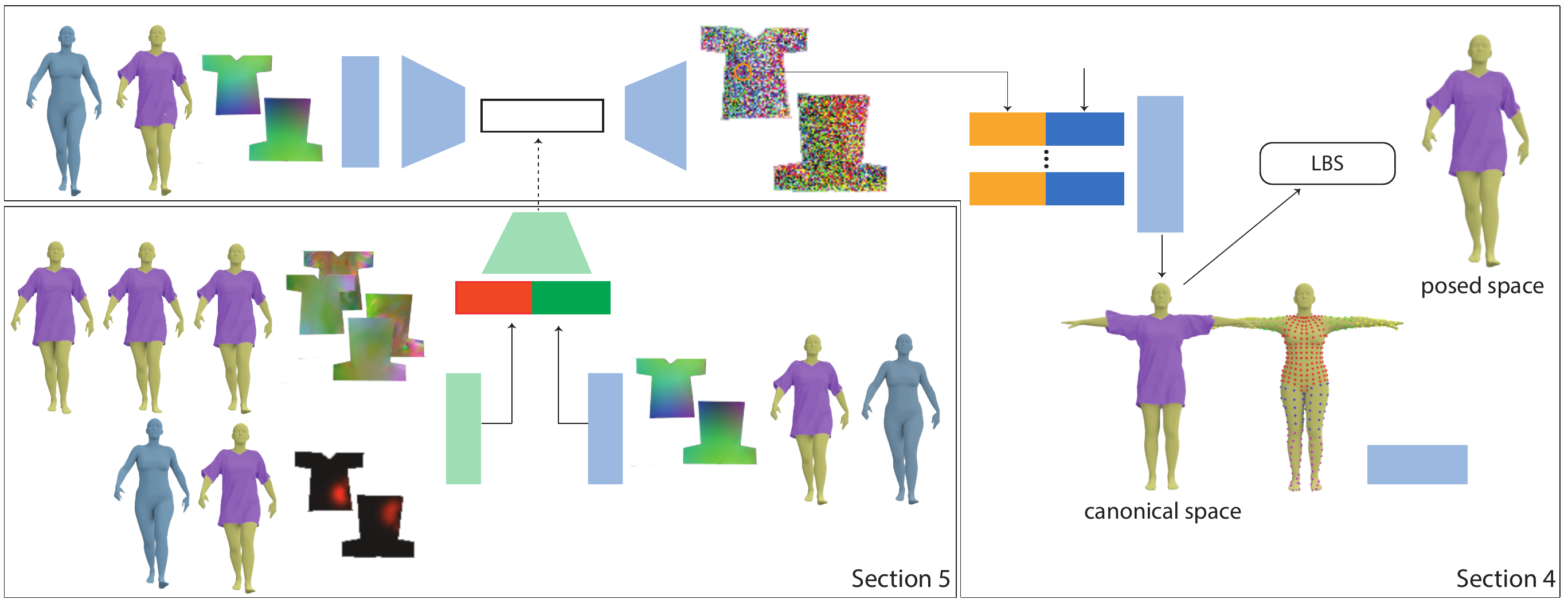}
   \put(1,36){
      \color{black} $B_{t}$
      }
     \put(7,36){
      \color{black} $G_{t}$
      }
      \put(13,36){
      \color{black} $M^P_{t}$
      }
      \put(1,23.5){
      \color{black} $G_{t-3}$
      }
      \put(7,23.5){
      \color{black} $G_{t-2}$
      }
      \put(13,23.5){
      \color{black} $G_{t-1}$
      }
      \put(6,1){
      \color{black} $B_{t}$
      }
      \put(11,1){
      \color{black} $G_{t-1}$
      }
      \put(22,1){
      \color{black} $C_{t}$
      }
      \put(33.5,30.5){
      \color{black} $Z_{t}$
      }
      \put(40.5,30.5){
      \color{black} $\mathcal D$
      }
      \put(32,22){
      \color{black} $\mathcal E^{Dyn}$
      }
      \put(72.7,27.5){
      \color{black} $\mathcal R$
      }
      \put(89.3,8.5){
      \color{black} $\rho_{l}$
      }
      \put(52,36){
      \color{black} $M^{\xi}_t$
      }
      \put(64,34.5){
      \color{black} $\xi^i_t$
      }
      \put(68,34.5){
      \color{black} $u^i$
      }
      \put(25.5,30.5){
      \color{black} $\mathcal E^{Sta}$
      }
      \put(22,30.5){
      \color{black} $\mathcal S$
      }
      \put(71.5,21.5){
      \color{black} $d_t^i$
      }
      \put(80,23){
      \color{black} $w_t^i$
      }
      \put(23,23){
      \color{black} $M^{V,\dot{V}}_{t-1}$
      }
      \put(28.5,10.5){
      \color{black} $\mathcal C$
      }
      \put(37.5,10.5){
      \color{black} $\mathcal S$
      }
      \put(42,16){
      \color{black} $M^P_{t-1}$
      }
      \put(30,16){
      \color{black} $E_{t}$
      }
      \put(50,18){
      \color{black} $G_{t-1}$
      }
      \put(56,18){
      \color{black} $B_{t-1}$
      }
      \put(79,8){
      \color{black} $B_0$
      }
      \put(70,8){
      \color{black} $\hat{G}_t$
      }
      \put(95,36){
      \color{black} $G_t$
      }
  \end{overpic}
  \caption{
  \textbf{Deep dynamic garment architecture. } Our approach first learns a compact generative space of plausible garment deformations. We achieve this by encoding a garment geometry $G_t$ represented as relative to the underlying body $B_t$ (i.e., $M^P_t$) to a latent code $Z_t$ using $\mathcal E ^{Sta}$. A decoder $\mathcal D$ then predicts a geometry feature map $M^{\xi}_t$. We sample $M^{\xi}_t$ to obtain per-vertex geometry features $\xi^i_t$. These features $\xi^i_t$ along with the vertex UV coordinates $u^i$ are provided to an MLP, $\mathcal R$, to predict per-vertex canonical space displacements $d^i_t$. We assign each vertex a skinning weight $w_t^i$ based on its proximity to the underlying body seed points weighted by a per-body part learnable kernel radius $\rho_l$. Once a generative space of garment deformations are learned, we train a dynamic-aware encoder $\mathcal E^{Dyn}$. We provide the previous garment geometry ($M^P_{t-1}$), the garment velocity and acceleration ($M^{V,\dot{V}}_{t-1}$), and the interaction between the body and the garment ($C_t$) as input. The encoder $\mathcal E^{Dyn}$ maps these inputs to a latent code $Z_t$ in the learned deformation space which is then used to decode the current garment geometry $G_t$. Blocks denoted in blue are pre-trained and kept fixed when training the blocks in green.}
  
  \label{fig:network}
\end{figure*}
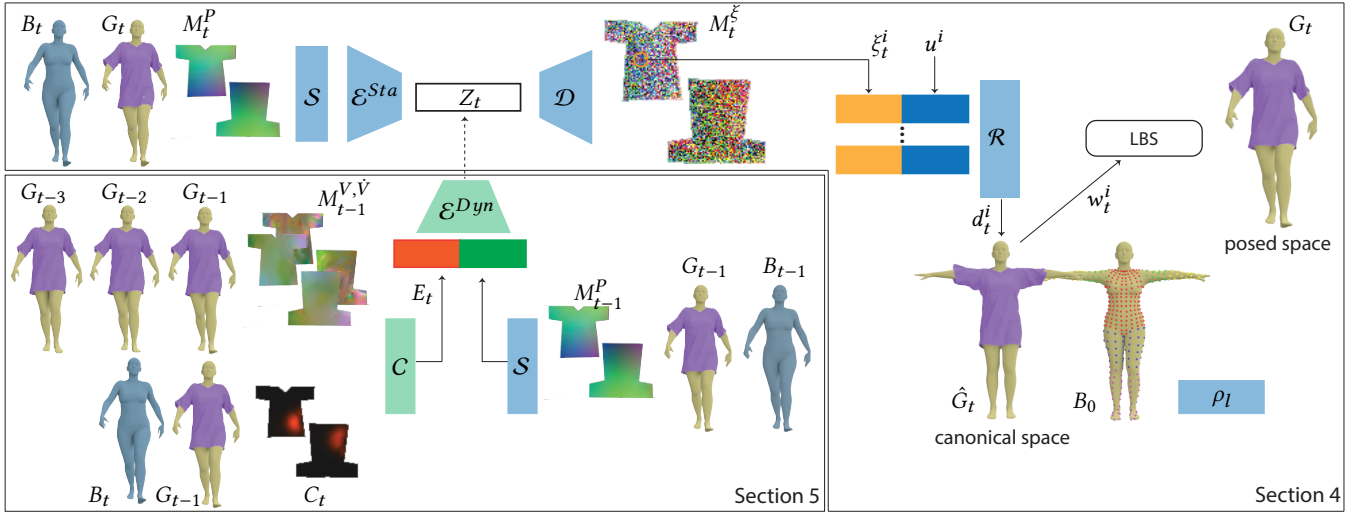

 Given the body state at a current frame  $B_t$, the relative transformation of each body vertex $b^j$ with respect to the canonical pose $B_0$ is represented by a known rotation and a translation $\{R_t^j,T_t^j\}$. If we denote the per-vertex local displacements for each garment vertex in the current frame as $d_t^i$ and the linear blending weights with respect to each body vertex as $w_t^i:=\{w_t^{i j}\}_j$, then the final position of the vertex, $g_t^i$, can be expressed as: 
 \begin{eqnarray}
    g_t^i &=& \sum_{j\in J} w_t^{i j}(R_t^j(\hat{g}_t^i-b_0^j)+b_0^j+T_t^j), \nonumber \\
     \hat{g}_t^i & =& g_0^i+H_0^i d_t^i.
 \end{eqnarray}
 We represent the local displacements in a per-vertex local coordinate frame using the normal and tangent vectors defined at $g_0^i$ and $H_0^i$ denotes such local coordinate frames. Our experiments show that displacements represented in such local frames lead to more stable training since otherwise quite large displacement values need to be predicted based on the underlying motion. Further, for efficiency, instead of utilizing all the body vertices, we regularly sample a set of seed points $\{b^j\}_{j\in J}$ on the body surface, as shown in Figure~\ref{fig:network}. {We sample such points regularly in the body uv space which we observe to be distributed evenly on the surfaces of the respective body parts using geodesic-based farthest point sampling.}
 
 
\subsection{Generative garment deformations} \label{subsec:garment geometry prediction}
It is challenging to learn a data driven model that can approximate the large space of body motion and corresponding garment configurations from a small set of training data. Hence, we seek an effective generative model that can represent the space of plausible garment deformations. We achieve this goal by utilizing a regularized autoencoder that can learn a mapping from $G_t$, the garment geometry at a particular frame $t$, to the garment deformation parameters, namely the displacements, $D_t$, in the canonical pose as well as the blending weights $W_t$ through a compact latent space. Instead of representing $G_t$ in absolute coordinates, we encode it relative to the underlying body. This relative encoding results in an input space with lower variation making it easier to learn and generalize to unseen body shapes (see results in Section \ref{sec:exp}). Specifically, given the garment geometry $G_t$ and the body $B_t$, we define $P_t:=\{p_t^i\}$ where $p_t^i:=[\vec{p}_t^{i j}]_{j \in J}=[g^i_{t}-b_t^j]_{j\in J}$ encodes the relative position between a garment vertex $g^i_{t}$ and a set of seed points $\{b^j_t\}_{j\in J}$ sampled on the body. We encode the position of each vertex with respect to each seed point. 

We record the body-relative garment descriptor $P_t$ in a 2D map corresponding to the UV space of the garment as $M^P_t \in \mathbb{R}^{w \times h \times (3N)}$ where $(w,h)$ are the dimensions of the pre-defined UV map and $N$ is the number of seed points sampled on the body (we set $w=h=128$ and $N=581$ in our experiments). Encoding the information in the UV space enables to exploit the locality of the 2D convolutions and helps to capture the neighborhood information~\cite{ma2021scale}. We first use a few convolutional layers $S$ to map $M^P_t$ to a predefined feature dimension $S(M^P_t) \in \mathbb{R}^{w \times h \times 128}$. Then, we employ a feature map encoder, $\mathcal{E}^{Sta}$, composed of 2D convolutions and a linear perceptron layer to output a latent vector $Z_t \in \mathbb{R}^{64}$:
\begin{equation}
    Z_t = \mathcal{E}^{Sta}(\mathcal{S}(M^P_t)).
\end{equation}
We expect $Z_t$ to encode plausible garment deformations and hence utilize it in subsequent stages to decode plausible deformation parameters, $(D_t, W_t)$.

\paragraph{Displacement prediction. }
Given $Z_t$, we use a decoder $\mathcal{D}$ with an architecture symmetric to the encoder $\mathcal{E}^{Sta}$ to decode a geometry feature map $M^{\xi}_t \in \mathbb{R}^{w \times h \times 128}$:
\begin{equation}
    M^{\xi}_t = \mathcal{D}(Z_t).
\end{equation}
Finally, we use a multi-linear perceptron (MLP) $\mathcal{R}$ to predict per-vertex local displacements $d_t^i$:
\begin{equation}
    d_t^i = \mathcal{R}(\xi^i_t, u^i),
\end{equation}
where $\xi^i_t$ is the per-vertex geometry feature sampled by bilinear interpolation from the geometry feature map $M^{\xi}_t$ based on its UV coordinate $u^i$. We illustrate the network architecture in detail in Figure~\ref{fig:network}.

\paragraph{Linear blending weight prediction. }
Given the local displacements $d^i_t$, we can compute each garment vertex position in the canonical space, $\hat{g}_t^i$. Then, we compute the per-frame linear blending weights $w_t^{ij}$ based on the distance between the garment vertex $\hat{g}_t^i$ and the body seed point $b^j_0$ in the canonical pose. Specifically, we set the linear blending weight as:
\begin{linenomath}
\begin{equation}
    w_t^{ij} = \frac{s_t^{ij}}{\sum_{k\in J}{s_t^{ik}}}, \quad \text{and} \quad 
    s_t^{ij} = exp
    \left(-\frac{\|\hat{g}_t^i-b^j_0\|^2}{2\rho_{l(j)}^2}\right),
\end{equation}
\end{linenomath}
where $\rho_{l(j)}$ is a learnable kernel radius assigned to the seed point based on the body part it belongs to. In other words, we learn a kernel radius for each body part (i.e., upper body, fore-arm, rear-arm, thigh and calf as shown in Figure~\ref{fig:network}). We observe that using such a part-based kernel function to predict blending weights provides a good trade off in terms of robustness and capturing dynamics compared to using a fixed or unconstrained blending weights, as we experiment in Section~\ref{sec:exp}. The final garment vertex position $g_t^i$ is computed by linear skinning using the predicted skinning weights $\{w_t^{ij}\}$, as described in Section~\ref{Sec: repre_garment_deformation}.

\paragraph{Training details. }
We train our variational autoencoder to jointly learn the parameters of $S$, $\mathcal{E}^{Sta}$, $\mathcal{D}$, $\mathcal{R}$, and the kernel radius $\{\rho_l\}$.

We enforce the predicted garment geometry $G_t$ (after applying local displacements $D_t$ and skinning with $W_t$) to be similar to the ground truth $G^*_t$ with the reconstruction loss:
\begin{equation}
    L_{rec} = \|G_t-G_t^{*}\|_1 + \|\Delta G_t - \Delta G_t^*\|_1,
\end{equation}
where $\Delta$ is the mesh Laplacian operator. 

A common practice in training variational autoencoders to ensure a compact and smooth latent space is to enforce a Gaussian prior (e.g., a normal distribution $\mathcal{N}(0, 1)$) on the latent codes using a KL-divergence score~\cite{Kingma2014}. Since, we have only a limited number of training samples (e.g., $300$ samples in our experiments), we observe that this does not work well in our setup. Instead, we apply a regularization term to ensure any latent code $z_t$ is within the normal distribution $\mathcal{N}(0, 1)$:
\begin{equation}
    L_{reg} = \left\|{\sum z_t^2}/{|Z|}-1\right\|_1,
\end{equation}
where $|Z|=64$ is the dimension of the latent codes.

Inspired by Santesteban et al.~\shortcite{santesteban2021self}, to ensure that any sampled latent code results in a plausible garment deformation, during training we randomly sample a latent code $Z_r$ within the normal distribution, and decode the corresponding garment geometry $\hat{G_r}$ in the canonical space. We enforce the edge length of $\hat{G_r}$ to be close to that of the canonical garment $G_0$ using the loss, 
\begin{equation}
   L_{rand} = \|Edge(\hat{G}_r)-Edge(G_0)\|_1.
\end{equation}
    
We pre-train a signed distance prediction network \cite{icml2020_2086} ($SDF_0$) to approximate the signed distance of any garment vertex $\hat{g}^i$ to the canonical body $B_0$. We regularize the displacement prediction by enforcing collision-free geometry in the canonical space on both the training samples $\hat{G}_t$ and the randomly sampled deformed garments $\hat{G}_r$:
\begin{equation}
    L_{SDF_0} = \max (\epsilon-SDF_0(\hat{G}_t), 0) + \max (\epsilon-SDF_0(\hat{G}_r), 0).
\end{equation}
The final loss function is a linear combination of the terms:
\begin{equation}
    L=L_{rec} + L_{rand} + \lambda_1 L_{SDF_0}. + \lambda_2 L_{reg}.
\end{equation}
In our experiments, we set $\lambda_1 = 100$, and $\lambda_2 = 0.001$.

\section{Dynamic-aware Garment Deformations} \label{sec:dynamic-aware encoding}
Once we learn a latent space of plausible garment deformations, our next goal is to predict the garment deformation at a particular frame $t$, as a latent code in this space, by taking into account the underlying body motion and the garment dynamics. Specifically, we seek an encoder that can map the previous state of the garment and the body to a latent code $z_t$, which can then be decoded to obtain the garment geometry $G_t$ in the current frame. Next, we describe the inputs of this encoder and the training procedure.

\paragraph{Dynamic-aware inputs.}
One of the factors that affects the garment deformation is its inertia, the tendency to preserve its state. In order to capture this behaviour, we record the velocity $V_{t-1}$ and the acceleration $\dot{V}_{t-1}$ of the garment in the previous time step in the garment UV space as a \emph{garment motion feature map}, $M^{V,\dot{V}}_{t-1}$. We provide $M^{V,\dot{V}}_{t-1}$ as one of the inputs to our dynamic-aware encoder.

Another important factor in how the garment deforms is its interaction with the underlying body. To capture this information, we introduce a per-vertex interaction feature for each vertex of the garment at the previous step, $g_{t-1}^i$, with respect to the body at the current step. 
Specifically, for a garment vertex $g_{t-1}^i$, we first compute if there is a collision with any of the body seed vertices $b_t^j$ by evaluating the signed distance from $g_{t-1}^i$ to the tangent plane of $b_t^j$ defined by its unit normal $n_t^{bj}$:
\begin{equation}
    q_t^{ij} := (g_{t-1}^i-b_t^j) \cdot n_t^{bj}.
\end{equation}
In case of a potential collision, there is an interaction force between the body and the garment vertices. We characterize the magnitude of the force as the penetration amount using a $Relu(*)$ function:
\begin{equation}
    Relu(-q_t^{ij})= \left\{ \begin{array}{rcl}
         -q_t^{ij} & \mbox{if} & q_t^{ij} < 0 \\
         0 & \mbox{if} & q_t^{ij} \ge 0. 
    \end{array} 
    \right.
\end{equation}

We use the normal of the body vertex $n_t^{bj}$ as the interaction force direction. Finally, we weight each interaction force by the distance between the garment vertex $g_{t-1}^i$ and the body seed point $b^j_{t-1}$: $a_t^{ij}:= exp(-\|g_{t-1}^i-b^j_{t-1}\|^2/(2\sigma^2))$. Hence, the resulting interaction force ${f}_t^{ij}$ between the garment vertex $g_{t-1}^i$ and the body vertex $b_t^j$ is formulated as:
\begin{equation}
    {f}^{ij}_t :=a_t^{ij} Relu(-q_t^{ij})n_t^{bj}.
\end{equation}
We record such interaction forces in the garment UV space, $C_t := \{c_t^i\}:=\{[{f}_t^{ij}]_{j\in J}\}$ (as shown in Figure~\ref{fig:network}) and provide as additional input to the dynamic-aware encoder.

\paragraph{Dynamic-aware encoder. }
Given the garment motion feature map, $M^{V,\dot{V}}_{t-1}$, and the interaction feature map, $C_t$, we first use a 2D convolutional encoder $\mathcal{C}$ to implicitly encode the relative state of the garment with respect to the body: 
\begin{equation}
    E_t := \mathcal{C}(M^{V,\dot{V}}_{t-1}, C_t).
\end{equation}

As described in Section \ref{subsec:garment geometry prediction}, we also encode the garment geometry in the previous frame as $\mathcal{S}(M^P_{t-1})$. We then introduce a dynamic encoder $\mathcal{E}^{Dyn}$ that maps $\mathcal{S}(M^P_{t-1})$ concatenated with $E_t$ into the learned latent space of garment deformations:
\begin{equation}
    Z_t := \mathcal{E}^{Dyn}(S(M^P_{t-1}), E_t).
\end{equation}
We expect $E_t$ to encode the change in the relative state of the garment with respect to the underlying body. In specific cases where both the body and the garment preserve their relative states, we can expect $E_t$ to be zero. We use this observation to introduce an additional constraint when training the dynamic encoder $\mathcal{E}^{Dyn}$. Specifically, we generate a new \emph{virtual} training sample by providing the inputs $S(M^P_{t-1})$ and $E_t=0$, we expect the latent code of this virtual sample, $Z^v_{t-1} = \mathcal{E}^{Dyn}(S(M^P_{t-1}), 0)$ to be the same as the latent code of the previous frame.

Once we obtain the latent codes $Z_t$ and $Z^v_{t-1}$, we use the pre-trained decoder $\mathcal{D}$ and the MLP $\mathcal{R}$ introduced in Section \ref{subsec:garment geometry prediction}, to predict the canonical space displacements $D_t$ and $D^v_{t-1}$ and compute the linear skinning weights with the learned kernel radius $\{\rho_l\}$ to generate the final posed garment geometries $G_t$ and $G^v_{t-1}$.

As shown in Figure \ref{fig:network}, with pre-trained blocks  $\mathcal{S}, \mathcal{D},  \mathcal{R}, \{\rho_l\}$ fixed, we train $\mathcal{C}$ and $\mathcal{E}^{Dyn}$ with the following loss function:
\begin{eqnarray}
    Loss &=& L_{geo} + L_{Z} + \lambda_2 L_{reg} \nonumber \\
    L_{geo} &=& \|G_t-G^*_t\|_1 + \|G^v_{t-1}-G^*_{t-1}\|_1 \nonumber \\
     L_{Z} &=& \|Z_t-Z^*_{t}\|_1 + \|Z^v_{t-1} - Z^*_{t-1}\|_1 \nonumber \\
     L_{reg} &=&  \|\frac{\sum z_t^2}{|Z|}-1\|_1+\|\frac{\sum{{z^v_{t-1}}^2}}{|Z|}-1\|_1,
\end{eqnarray}
where $G_t^*, G_{t-1}^*$ are the ground truth garment geometry for the current and previous frames, $Z^*_t, Z^*_{t-1}$ are the corresponding latent codes obtained by running the pre-trained $\mathcal{E}^{Sta}$ (Section \ref{subsec:garment geometry prediction}) on the ground truth garment geometries. We set $\lambda_2 = 0.001$ in our experiments.

\begin{figure*}
  \includegraphics[width=\textwidth]{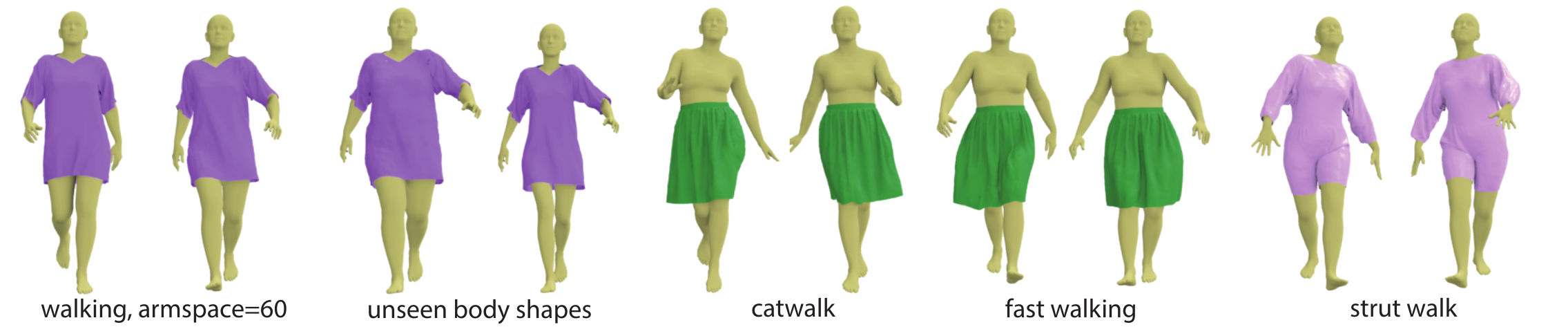}
  \caption{
  \textbf{Generalization to body shape and walking style. }
  We train our network on a walking sequence of $300$ frames on a fixed body shape and test on walking motion with different character armspace settings, different styles of walking, and different body shapes.}
  \label{fig:results}
\end{figure*}

\subsection{Explicit collision handling}
While we enforce to obtain collision free garment geometries in the canonical pose, we observe that for unseen motion sequences, occasional collisions remain unsolved in the posed space. To address this challenge, we present a novel collision resolving stage, at run time, that optimizes for a residual displacement map.


As described previously, our method predicts a per-vertex local displacement $d_t^i$ and blending weights $w_t^i$ to compute the deformed garment $G_t$. We introduce residual local displacements $\theta_i$ to resolve any remaining collisions. Specifically, we obtain the final garment geometry as:
\begin{eqnarray}
     \check{g}_t^i&=&g_0^i+H_0^i (d_t^i+\theta_i)=\hat{g}_t^i+H_0^i \theta_i,  \nonumber\\ \nonumber
      \tilde{g}_t^i&=&\sum_{j\in J} w_t^{i j}(R_t^j(\check{g}_t^i-b_0^j)+b_0^j+T_t^j).
 \end{eqnarray}
 In practice, we optimize for a 2D residual displacement map $\Theta$, defined in the UV space of the garment, and obtain $\theta_i$ by bi-linear interpolation which ensures smoothness.

For every garment vertex $g_t^i$ in the garment prediction $G_t$, we first get its closest body vertex $b_t^k$. We then compute the signed distance between $g_t^i$ and the tangent plane defined on $b_t^k$ by the vertex normal $n_t^{bk}$: 
\begin{equation}
    o_t^{ik} := (g_t^i-b_t^k)\cdot n_t^{bk}.
\end{equation}
In a collision-free case, we expect $o_t^{ik}$ to be positive, which means $Relu(-o_t^{ik}) = 0$. We define a collision loss for the collision detection:
\begin{equation}
    L_{collision} = \sum_i {Relu(-o_t^{ik})}.
\end{equation}
When predicting a dynamic garment sequence in a roll-out fashion (i.e., treat the prediction in the previous frame as inputs for the current frame), we set a threshold $\epsilon$ for the collision loss $L_{collision}$. When collisions occur, i.e., $L_{collision} > \epsilon$, we optimize for the displacement map $\Theta$ that minimizes the following objective function:
\begin{equation}
    L_{\Theta}=\|\tilde{G}_t-G_t\|_1 + \|\Delta \tilde{G}_t - \Delta G_t\|_1 + \lambda_3 L_{collision},
\end{equation}
where $\Delta$ is the Laplacian operator, and $\lambda_3 = 100$.

Once we optimize for $\Theta$ for a frame $t$, the {collision-resolved} garment geometry that incorporates $\Theta$ is provided as input to the network for the next frame. 

\section{Results and experiments}
\label{sec:exp}

\subsection{Data generation}
We rig and animate a body shape sampled from SMPL~\cite{SMPL:2015} via Mixamo\footnote{https://www.mixamo.com/} to generate a training walking motion sequence of $300$ frames. We set the arm space setting for the character as $75$ during training. Next, we run a physically-based simulation using Marvelous Designer\footnote{https://marvelousdesigner.com/} to generate the ground truth garment mesh sequence. We model three garment outfits: a short sleeve \emph{tshirt} ($18902$ vertices), a \emph{bodysuit} ($21904$ vertices), and a pleated \emph{skirt} ($17678$ vertices). We train our network in a fully supervised manner using the simulation output as ground truth. 

\begin{figure}[h!]
   \includegraphics[width=\columnwidth]{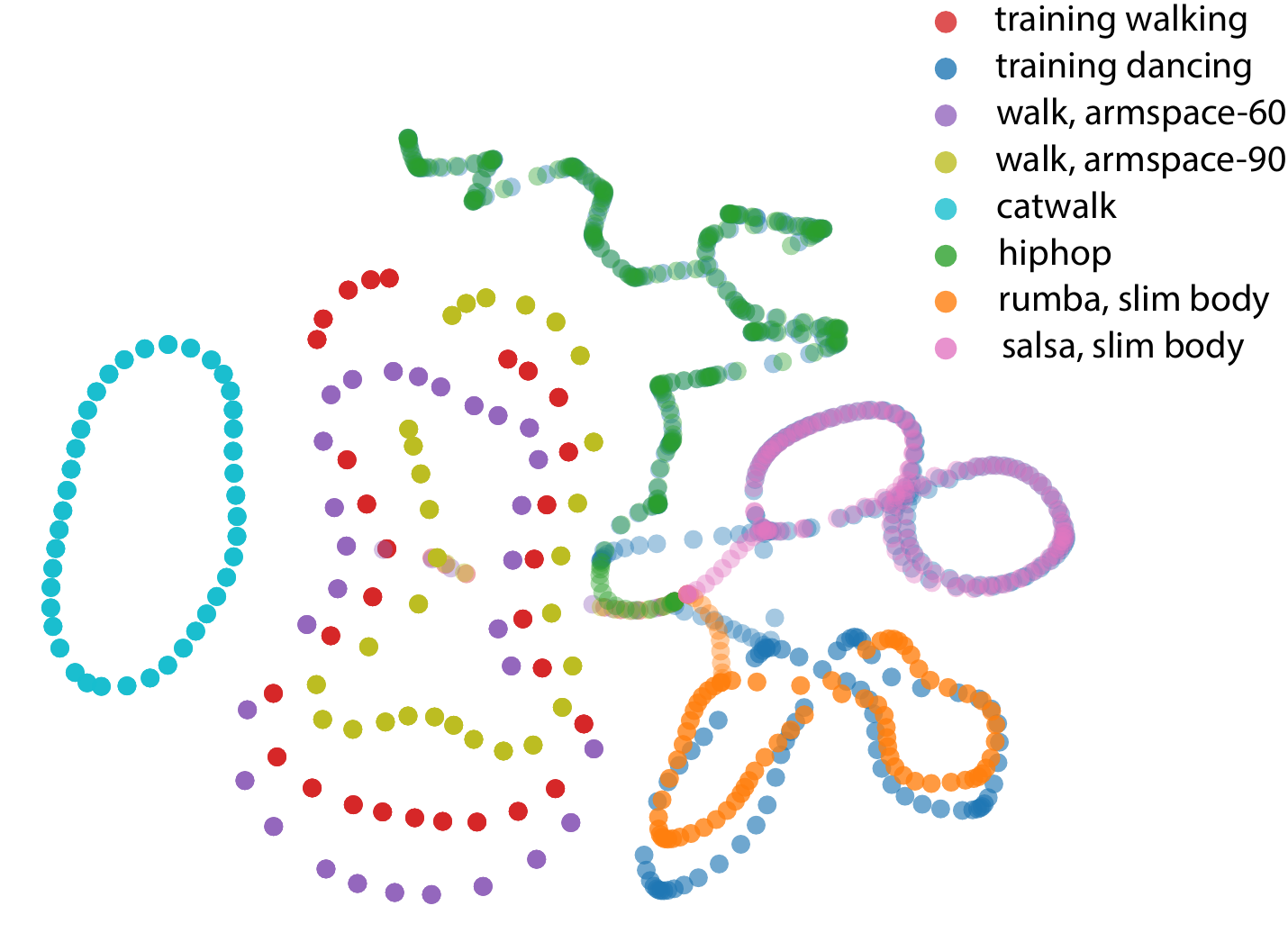}
   \centering
   \caption{
   {We visualize the distribution of the training and testing motion sequences via t-SNE~\cite{maaten2008visualizing}. The testing motions of armspace-60 and armspace-90 are close to the training walking motion; hiphop, rumba, and salsa motions are within the distribution of the training dancing motion; and catwalk is away from the distribution of either training walking or training dancing motions.}
  }
   \label{fig:tSNE}
 \end{figure}

\subsection{Implementation details}
{We use $4$ layers of 2D convolutional layers to project the recorded body-related garment geometry map ($M_t^P$) to a static 128D feature map ($S(M^P_t)$). Our static encoder ($\mathcal{E}^{Sta}$) consists of $6$ 2D convolutional layers to gradually downsample the resolution of the feature map from $128 \times 128$ to $2 \times 2$ and increase the feature dimension to $128$, $256$, $512$, $1024$, $1024$, $1024$ respectively. We use instance normalization and leakyReLU for all layers. We flatten the encoded feature map ($2 \times 2 \times 1024$) as a 4096D vector and linearly project it into a compact and regularized 64D latent space ($Z_t$) by a fully connected layer. The feature map decoder ($\mathcal{D}$) reverses the encoder architecture symmetrically, along with one more transpose convolutional layer to upsample the feature map to size $256 \times 256$. Finally, $6$ layers of linear perceptrons ($\mathcal{R}$) decode the vertex-wise features sampled from the resulting feature map and predict the position of each vertex.

The dynamic encoder ($\mathcal{E}^{Dyn}$) has a similar architecture as the static encoder, except that it expects an input feature map with $256$ channels (static geometry feature map $S(M^P_{t-1})$ concatenated with the motion feature map $E_t$). The motion feature map $E_t$, with the same size as the static feature map (i.e., $128 \times 128 \times 128$), is generated by $6$ 2D convolutional layers ($\mathcal{C}$) which take as input the body-interaction features $C_t$, garment velocity and acceleration $M^{V,\dot{V}}_{t-1}$. 

During training, we use the AdamOptimizer with a learning rate starting from $1.0^{-4}$ and halve it every $50$ epochs. In our experiments, it took around $350$ epochs to converge. We set the batch size varying from $1$ to $4$, based on the amount of the garment vertices. Our method still works well when the bath size is $1$, since the MLP decoder ($\mathcal{R}$) predicts the position of each vertex given the vertex-wise features sampled from the resulting feature maps.}
 
\begin{table}[h!]
    \caption{{To quantitatively evaluate the generalization ability, we show the average percentage of garment vertices inter-penetrating the body meshes across different testing motions produced by the two networks trained with walking motion sequence and dancing motion sequence respectively.}}
    \begin{tabular}{l||c|c}
        \diagbox{testing}{training} & walking & dancing\\
        \hline
        \hline
        walk, armspace-60 & $\textbf{0.15\%}$ & $0.73\%$ \\
        \hline
        walk, armspace-90 & $\textbf{0.01\%}$ & $0.13\%$ \\
        \hline
        catwalk & $\textbf{0.19\%}$ & $0.58\%$ \\
        \hline
        rumba, slim body & $0.11\%$ & $\textbf{0.08\%}$ \\
        \hline
        salsa, slim body & $0.22\%$ & $\textbf{0.10\%}$ \\
        \hline
        hiphop & $0.34\%$ & $\textbf{0.12\%}$ \\
        
    \end{tabular}
    \label{tab:collision}
\end{table}

\subsection{Results and evaluation}
\paragraph{Generalization.}
For all garment types, we train our network on a relatively short walking sequence of $\emph{300}$ frames simulated on a \emph{fixed} body shape. Once trained, we test our method on different styles of walking motion by changing the character arm space setting\footnote{This setting provided by Mixamo changes how the motion is retargeted to the character resulting in a different motion style.}, changing the speed of the motion, applying different types of walking (e.g., catwalk, strut walk), and changing the body shape. {We also further evaluate the generalization capability of our network by testing on challenging dancing motions (including salsa swing, rumba swing, hiphop).} {In order to more systematically analyze the generalization behaviour, we train a separate network with another short dancing sequence of \emph{$495$} frames simulated on the same \emph{fixed} body shape and test the trained network on the same testing motion sequences. In Figure \ref{fig:tSNE}, we visualize the distribution of both the training and testing motion sequences via t-SNE~\cite{van2008visualizing} computed over the 3D body seed points sampled on the dynamic body sequences.} 

We show qualitative results in Figure~\ref{fig:results} and the supplementary video. The use of a generative latent space of garment deformations helps to achieve generalization across different motion styles. 
{In Table~\ref{tab:collision}, we provide a quantitative comparison of the results across different testing sequences obtained by networks trained with walking and dancing motions respectively. 
As expected, our network shows better generalization for testing motion sequences which are distributed more closely to the training data than the unseen motions with significantly different data distribution. We observe that our network trained with normal walking sequence produces reasonable output for the unseen salsa motion but also lacks dynamics and suffers from collisions when tested on the hiphop sequence. When the network is trained on a similar dancing sequence, such artifacts are reduced.} 
Additionally, since we represent the garment geometry relative to the underlying body shape, our method also has plausible generalization ability for unseen body shapes.

\begin{figure}[h!]
  \includegraphics[width=\columnwidth]{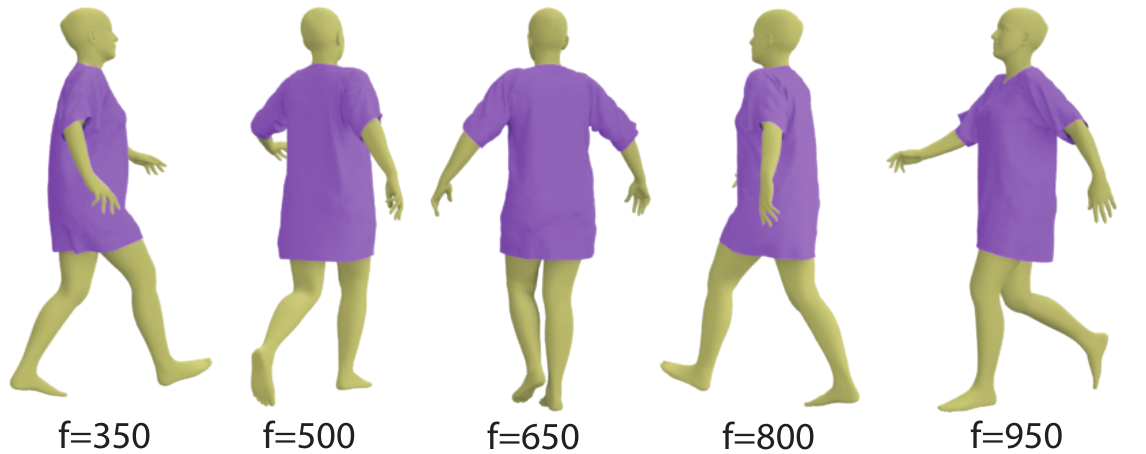}
  \caption{
 \textbf{Robustness under long roll-out. }
  Predictions of our network are utilized as input in subsequent frames and enable iterative roll-out predictions as long as 1000 frames. See supplemental video. }
  \label{fig:rollout}
\end{figure}

\begin{table}[b!]
    \small
    \caption{
    {To evaluate the stability of our network, we report the L2 error when predicting long motion sequences at test time by iterative roll-out prediction for more than a thousand frames.}}
    \begin{tabular}{c||c|c|c|c}
        \multirow{2}{*}{garment} & \multirow{2}{*}{motion} & 1-step & rollout-50 & rollout-1150\\
        & & L2 ($\times 10^{-2}$) & L2 ($\times 10^{-2}$) & L2 ($\times 10^{-2}$) \\
        \hline \hline
        T-shirt & armspace-90 & 0.59 & 0.83 & 0.82 \\
        \hline
        pleated-skirt & armspace-75 & 1.03 & 1.01 & 1.09 \\
        \hline
        bodysuit & armspace-90 & 0.89 & 1.08 & 1.06 \\
    \end{tabular}
    \label{tab:my_label}
\end{table}
\paragraph{Iterative roll-out prediction.} We test the stability of our network at predicting long motion sequences at test time by iterative roll-out prediction for more than a thousand frames. As shown in Figure~\ref{fig:rollout}, our network can utilize its predictions as input in the subsequent frames and produce reasonable predictions for long testing sequences. See also supplementary video. Table~\ref{tab:my_label} provides the error with respect to the ground truth for such roll-out predictions.

\begin{figure}[t!]
   \includegraphics[width=\columnwidth]{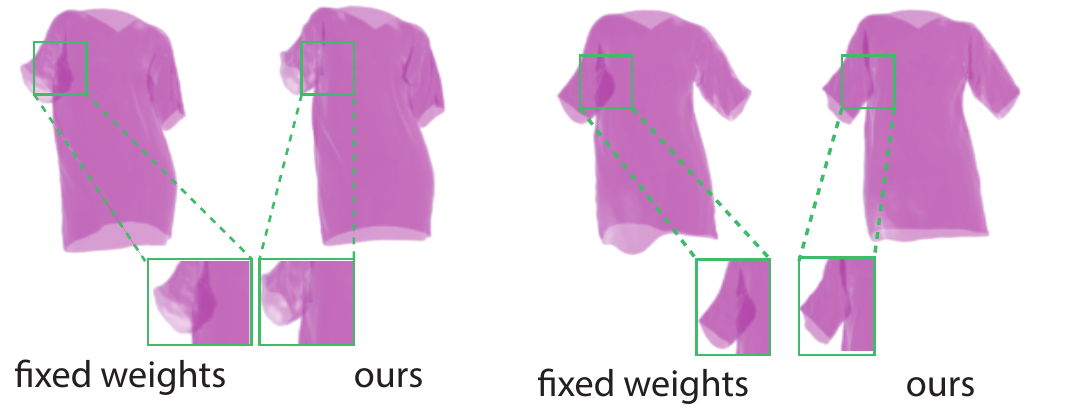}
   \caption{
  \textbf{Static versus Dynamic blending weights. }
   Using frame-dependent dynamic blending weights results in more plausible garment deformation compared to using a fixed set of blending weights.}
   \label{fig:weights}
 \end{figure}

\paragraph{Effect of dynamic blending weights.} Our network predicts garment deformation as a set of local displacements in the canonical configuration of the garment, and maps it to the global space using linear blend skinning. While it is possible to assign a fixed set of blending weights to each garment vertex (e.g., based on the proximity of the garment vertices to the body in the canonical pose), we find that predicting dynamically changing blending weights based on a learned per-part kernel radius provides more flexibility and results in more dynamic behavior. We show a comparison in Figure~\ref{fig:weights} and the supplementary video. For an unseen motion with arms closer to the body, the prediction with fixed weights introduces artefacts especially in the armpit regions.


\begin{figure}[b!]
   \includegraphics[width=\columnwidth]{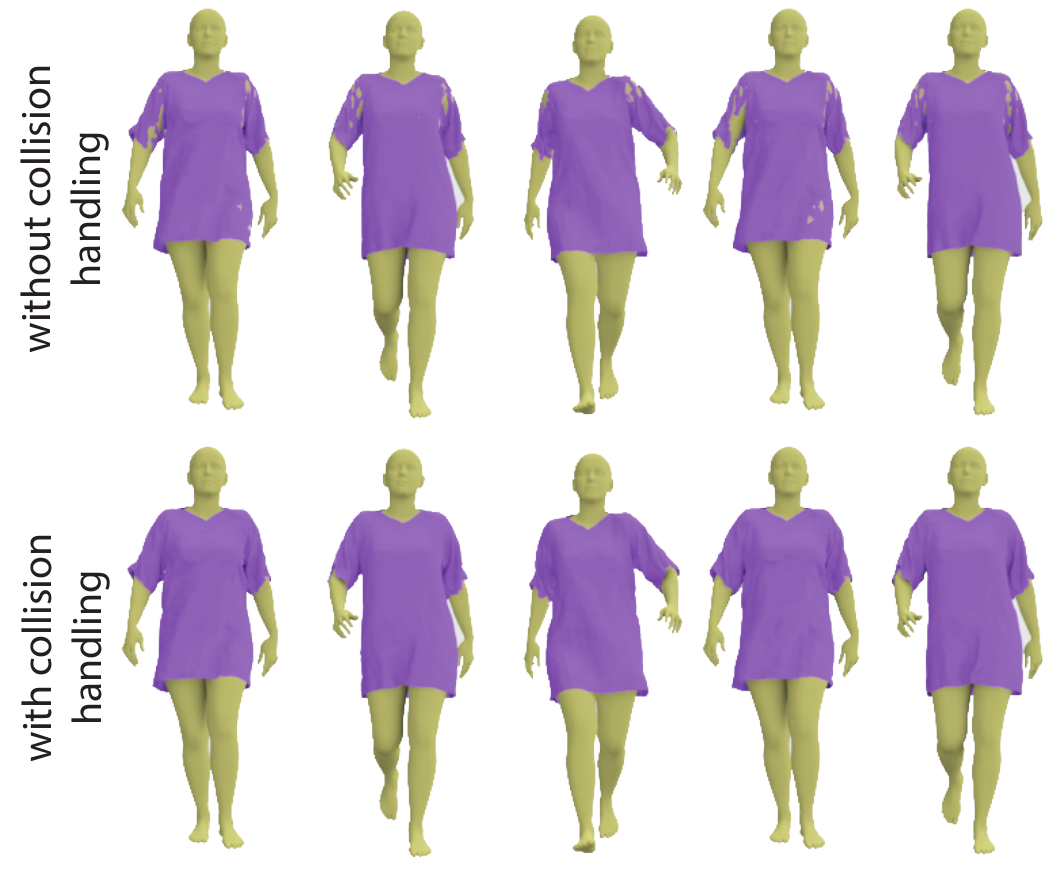}
   \caption{
   \textbf{Explicit collision handling. }
   We resolve any remaining body-garment collisions by fine tuning for for residual local displacements. 
   }
   \label{fig:collision}
 \end{figure}

\paragraph{Effect of explicit collision handling.} For unseen motion sequences, we observe that occasional collisions remain unsolved. We optimize for a residual displacement map to resolve such intersections between the garment and the body as shown in Figure~\ref{fig:collision}, this optimization is effective. Any residual local displacement is propagated to the subsequent frames implicitly by feeding the final garment as input back to the network. For the unseen motion sequence in Figure~\ref{fig:collision}, our collision handling optimization reduces the average percentage of the garment vertices inter-penetrating the body mesh from $7.01\%$ down to $0.15\%$. With our unoptimized Pytorch code, it takes about $0.26$ seconds to run a forward prediction without explicit collision handling ($0.20$ seconds to prepare the input maps, $0.01$ seconds for the inference with dynamic-aware encoder, and $0.05$ seconds to decode and compute garment vertex positions), and about $0.18$ seconds for one iteration of the residual map optimization ($0.08$ seconds to compute the closest body vertices and $0.10$ seconds to optimize the residual map). For the example in Figure~\ref{fig:collision}, it takes $110.25$ seconds in total to predict a sequence of garments with $200$ frames. 
 
\paragraph{Effect of training data length.} {Table~\ref{tab:SeqLen} shows the L2 error of roll-out prediction for an unseen walking motion sequence (armspace-90) by running the network trained with different length of the training walking data (armspace-75). When trained with $50$ frames, the network easily overfits to the seen motion resulting in a significant performance drop. Training the network with $900$ frames of cyclic motion, it makes the network statically prone to the seen motion (armspace 75). Thus when we test the network with the walking sequence of 90-armspace, the error increases for the short roll-out prediction. However, training with a long walking sequence improves the stability of the long roll-out prediction. In our experiments we use sequence of $300$ frames to achieve a good balance between the long and short roll-out prediction accuracy. We speculate that our generative architecture, encoding the 3D garment geometry as the local representation relative to the underlying body, enables our network to generalize across unseen motions, trained  with only a relative short sequence.
}

\begin{figure*}[h!]
   \includegraphics[width=\textwidth]{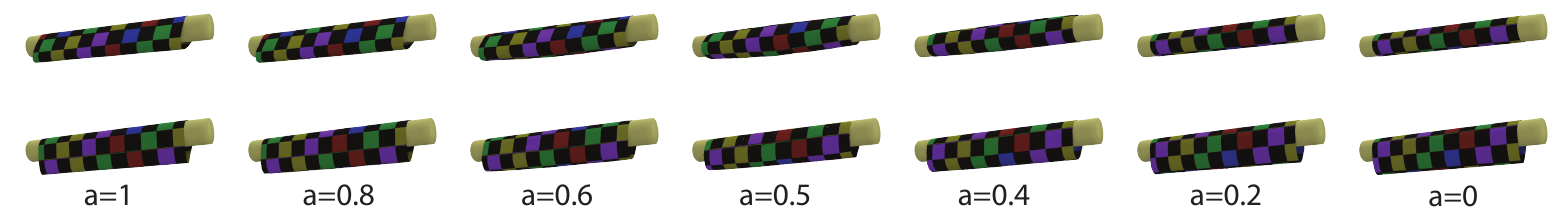}
   \caption{
   \textbf{Generative garment deformation space. }
   We train our network on a sequence of a sheet of cloth wrapped around a moving stick. We sample two latent codes and linearly interpolate between them using weights $a$ and $1-a$. 
   {The first row shows the interpolation between two different shapes. The second row shows how the cloth smoothly rolls up around the stick as can be observed from the checkerboard texture.}}
   \label{fig:latent}
 \end{figure*}
 
\begin{table}[b!]
    \caption{{To evaluate the effect of data length when training the network with the normal walking sequence (armspace-75), we report the L2 error of roll-out prediction at the long sequence of an unseen motion (armspace-90) by running the network trained with different length of the training data.}}
    \begin{tabular}{l||c|c|c}
        \multirow{2}{*}{\diagbox{training}{testing}} & 1-step & rollout-50 & rollout-1150 \\
         & L2 ($\times 10^{-2}$) & L2 ($\times 10^{-2}$) & L2 ($\times 10^{-2}$) \\
        \hline \hline
        with 50 frames & 0.63 & 1.10 & 1.84 \\
        \hline
        with 150 frames & \textbf{0.55} & \textbf{0.81} & 0.84 \\
        \hline
        with 300 frames & 0.59 & 0.83 & 0.82 \\
        \hline
        with 900 frames & 0.75 & 0.92 & \textbf{0.72} \\
    \end{tabular}
    
    \label{tab:SeqLen}
\end{table}
%


\begin{figure}[h!]
   \includegraphics[width=\columnwidth]{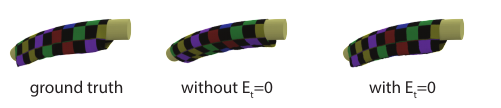}
   \caption{
   \textbf{Training of the dynamic-aware encoder. }
   When training the dynamic-aware encoder, we enforce the constraint that when both the body and the garment preserve their states ($E_t=0$), we should obtain the same latent code as in the previous frame. This results in capturing more accurate deformations.}
   \label{fig:dynamic}
 \end{figure}
 
\paragraph{Generative garment deformation space.} In order to show that we learn a compact and regularized space of garment deformations in the first part of our training, in Figure~\ref{fig:latent}, we visualize garment deformations that are obtained by interpolating between two randomly sampled latent codes. The resulting garments are plausible and change smoothly.

\paragraph{The training strategy for the dynamic-aware encoder.} When training our dynamic-aware encoder, for each sample in the training data, we introduce a constraint that if $E_t=0$ (i.e., both the body and the garment preserve their current states), we should obtain a latent code equal to the previous frame (see Section~\ref{sec:dynamic-aware encoding}). We evaluate the effectiveness of this constraint by learning the deformation of a sheet of cloth wrapped around a moving stick. We train our network with and without this constraint and observe that this constraint helps to capture more accurate deformations as shown in Figure~\ref{fig:dynamic} and the supplementary video.

\subsection{Baseline Comparisons}
\begin{figure*}
   \includegraphics[width=\textwidth]{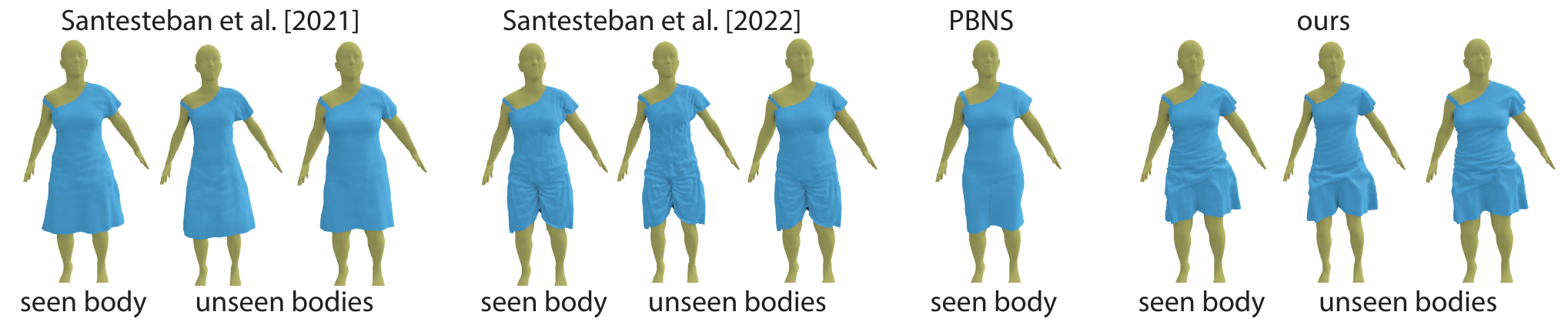}
   \caption{We compare our method to the works of Santesteban et al.~\shortcite{santesteban2021self,santesteban2022snug} and PBNS~\cite{Bertiche2021}. Our results generalize to unseen body shapes and capture more dynamic deformations.}
   \label{fig:comparison}
 \end{figure*}

We compare our method to recent learning-based garment deformation approaches on a dress example. Specifically, we compare to the works of Santestaban et al.~\shortcite{santesteban2021self,santesteban2022snug} which also learn dynamic deformations and PBNS~\cite{Bertiche2021} which provides an unsupervised setting for learning pose-dependent deformations using dynamic blending weights. As shown in Figure~\ref{fig:comparison} and the supplementary video, PBNS does not capture the dynamic deformations and results in a relatively stiff result. While the unsupervised method of Santesteban et al.~\shortcite{santesteban2022snug} uses GRUs to capture temporal information, the use of fixed blending weights results in inferior results where the dress appears to stick to the body. Santesteban et al.~\shortcite{santesteban2021self} produces more plausible results, but we observe that our method can capture more dynamic deformations.
{We speculate that two key aspects enable ours to produce more dynamic results. First, our network puts more attention on the local displacements, predicted from the body-relative garment geometry and body-garment interactions. In contrast, the baseline works utilize a simplified skeleton motion sequence as input. Second, our dynamic skinning weights are assigned to the body seed points, instead of the sparse skeleton joints, that capture more detailed deformations.}
Finally, while our method and the works of Santesteban et al.~\shortcite{santesteban2021self,santesteban2022snug} can be tested with unseen body shapes, PBNS is shape-specific. 
{Please note that, unlike Santesteban et al.~\shortcite{santesteban2021self,santesteban2022snug} who include multiple body shapes in the training data to generalize across unseen body shapes within the training distribution, we train on one fixed body shape. We enable body shape generalization by encoding the garment geometry relative to the underlying body, which exhibits lower variation across different body shapes.} 
We use the author provided trained models and the implementations for these comparisons.

\begin{figure}[b!]
   \includegraphics[width=\columnwidth]{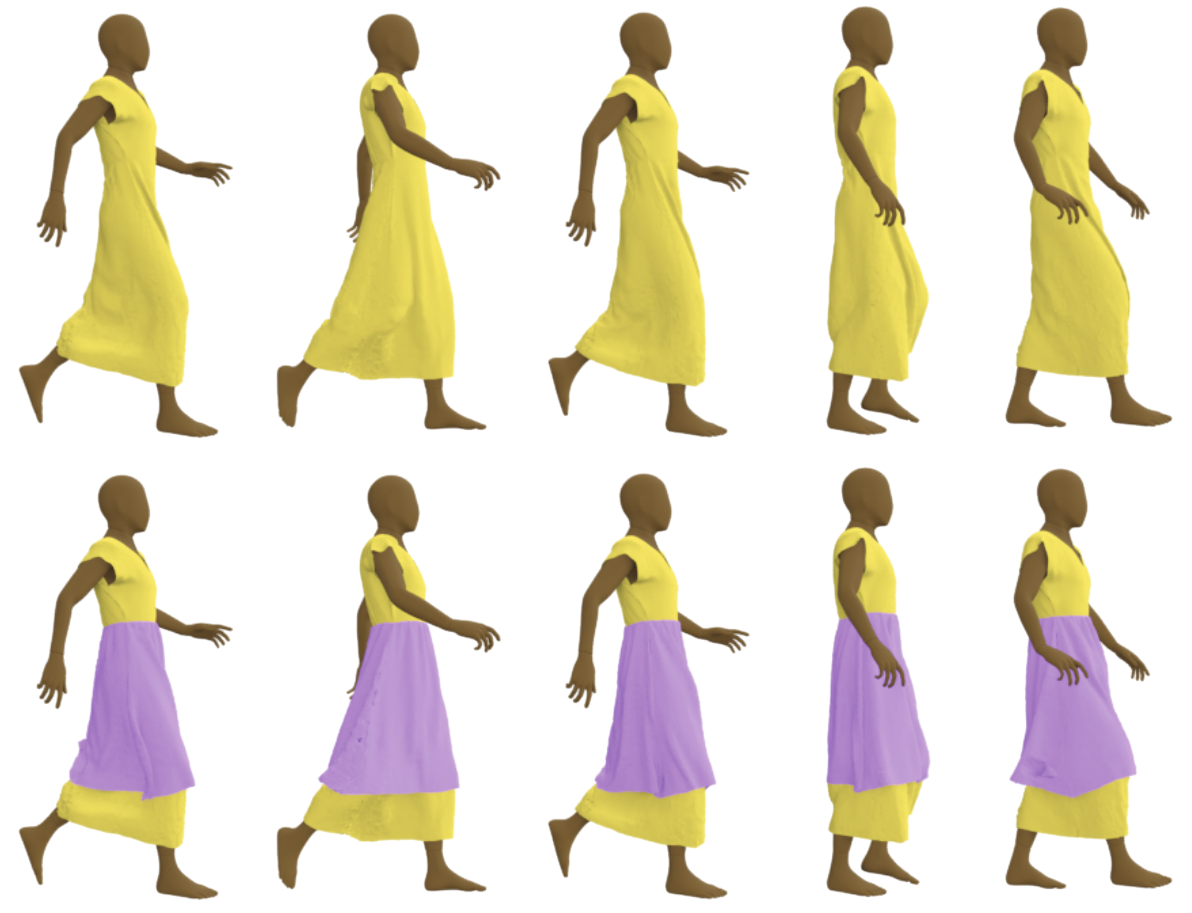}
   \caption{Our method can be extended to handle multi-layer garments. We train a network to predict how the yellow dress would deform based on the underlying body motion. Then, we train a second network to learn how the purple skirt deforms treating the yellow dress as the \emph{interaction body}. See supplementary video for results on seen and unseen motion. Note that this approach ignores the effect of the purple layer on the yellow layer.  }
   \label{fig:layers}
 \end{figure}
 
 \subsection{Extension to layered-garments}
 Our network encodes the garment geometry relative to the underlying body without making any assumptions about the body parameterization. This not only enables to generalize to unseen body shapes at test time but also paves the road to extend our work to handle layered garments. We provide an initial result towards this direction by evaluating our method on a two-layered garment as shown in Figure~\ref{fig:layers} and the supplementary video. Specifically, given the ground truth simulation data, we first train a network that learns how the yellow dress deforms given the underlying body motion. Next, we train another network for the purple skirt treating the yellow dress as the underlying body, i.e., we sample body seed points on the yellow dress. We assume that the canonical state of the yellow dress will stay fixed to make the training easy while in reality the canonical state of the yellow dress changes as well. Even with this simplifying assumption, we get reasonable deformation estimates for both layers. We believe that it is a very promising direction to further explore the opportunity to extend our method to handle multi-layer garment deformations.

\section{Conclusion}
We have presented a learning based method for capturing motion guided garment dynamics in 3D. The core of our method consists of a compact latent space of plausible garment deformations represented as canonical space displacements along with dynamically changing skinning weights. We then introduce a dynamic encoder that maps the previous states of the garment (geometry, velocity, and acceleration) as well as how it interacts with the body to this learned latent space to produce plausible dynamic deformations. While we enforce collision-free states in canonical pose, there may be remaining collisions between the garment and the body once posed. We resolve any such remaining collisions in a dynamic post processing step. Our method can predict the dynamic deformations of a garment for a long motion sequence by utilizing its output at the previous frame as input in the current frame. We demonstrate generalization to unseen motion types and body shapes. When compared to recent related work, our method captures more detailed dynamic deformations.

\begin{wrapfigure}{l}{0.2\columnwidth}
    \centering
    \includegraphics[width=0.2\columnwidth]{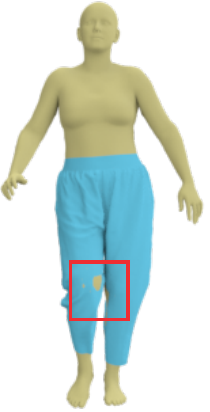}
\end{wrapfigure}
\paragraph{Limitations and future work.} While showing high quality 3D results, our method has limitations we would like to tackle in future
work. Our network can generalize to different styles of a particular motion, such as walking. However, generalizing across very different motion types is still challenging. 
While our test-time collision resolving optimization is effective, we observe that there still remains challenging scenarios when two different body parts come close to each other with loose garments as shown in the inset. 
Given the recent advances in implicit 3D representations, we would like to explore a more holistic approach of enforcing collision free geometry in both canonical and posed spaces during network training. This will be especially critical for handling multi-layer garments more effectively. 
Finally, training from real capture data is an exciting direction which will enable to learn the deformation properties of garments from observations without the need for manually setting material parameters and physically based simulation.

%

\begin{acks}
We would like to thank the anonymous reviewers for their
constructive comments; 
Igor Santesteban and Hugo Bertiche for the helping with the comparisons;
Mixamo for the motion sequences.
This work was partially supported by the ERC SmartGeometry grant, Marie Sklodowska-Curie grant 956585, and gifts from Adobe Research. 
\end{acks}

\bibliographystyle{ACM-Reference-Format}
\bibliography{main_bib}


\end{document}